\documentclass{bmvc2k}
\usepackage{subfiles}
\usepackage{graphicx, subfigure}
\usepackage[symbol]{footmisc}

\title{Learning Human Optical Flow}

\addauthor{Anurag Ranjan}{aranjan@tuebingen.mpg.de}{1}
\addauthor{Javier Romero$^{*,}$}{javier@amazon.com}{2}
\addauthor{Michael J. Black}{black@tuebingen.mpg.de}{1}

\addinstitution{
MPI for Intelligent Systems\\
T\"ubingen, Germany
}
\addinstitution{
Amazon Inc. \\
}

\footnotetext{*This work was done by JR while at MPI.}

\runninghead{Ranjan, Romero, Black}{Learning Human Optical Flow}

\begin{document}

\maketitle

\begin{abstract}
The optical flow of humans is well known to be useful for the analysis of human action.
Given this, we devise an optical flow algorithm specifically for human motion and show that it is superior to generic flow methods.
Designing a method by hand is impractical, so we develop a new training database of image sequences with ground truth optical flow.
For this we use a 3D model of the human body and motion capture data to synthesize realistic flow fields.
We then train a convolutional neural network to estimate human flow fields from pairs of images.
Since many applications in human motion analysis depend on speed, and we anticipate mobile applications, we base our method on SpyNet with several modifications.
We demonstrate that our trained network is more accurate than a wide range of top methods on held-out test data 
and that it generalizes well to real image sequences.
When combined with a person detector/tracker, the approach provides a full solution to the problem of 2D human flow estimation.
Both the code and the dataset are available for research.
\end{abstract}

\section{Introduction}








\begin{figure*}[t]
\centerline{
\includegraphics[width=\linewidth]{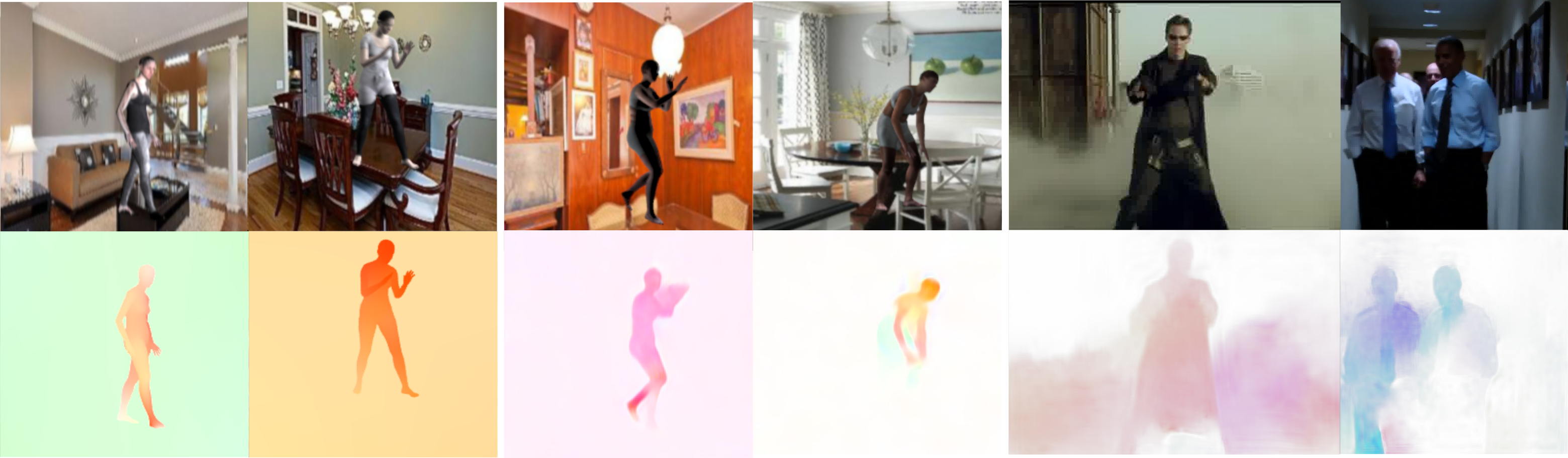}
}
\begin{flushleft}
\hspace{0.35in} \small{(a) Our dataset \hspace{0.48in} (b) Results on synthetic scenes \hspace{0.1in} (c) Results on real world scenes}
\end{flushleft}
\vspace{-0.1in}
\caption{(a) We simulate human motion in virtual world creating an extensive dataset with images (top row) and flow fields (bottom row); color coding from \cite{baker2011database}. (b) We train an existing deep network for human motion estimation  and show that it performs better when trained on our dataset and (c) generalizes to human motions in real world scenes.}
\label{fig:teaser}
\vspace{-0.7cm}
\end{figure*}

A significant fraction of videos on the Internet contain people moving \cite{scienceselfies} and the literature suggests that optical flow plays an important role in understanding human action \cite{Jhuang:ICCV:2013,soomro2012ucf101}.
Several action recognition datasets \cite{soomro2012ucf101,kuehne2011hmdb} contain human motion as a major component.
The 2D motion of humans in video, or {\em human flow}, is an important feature that provides a building block for
systems that can understand and interact with humans.
Human flow is useful for various applications including analyzing pedestrians in road sequences, motion-controlled gaming, activity recognition, human pose estimation system,  etc.

Despite this, optical flow has previously been treated as a generic, low-level, vision problem.
Given the importance of people, and the value of optical flow in understanding them, we develop a flow algorithm that is specifically tailored to humans and their motion.
Such motions are non-trivial since humans are complex, articulated, objects that vary in shape, size and appearance.
They move quickly, self occlude, and adopt a wide range of poses.



Our goal is to obtain more accurate 2D motion estimates for human bodies by training a flow algorithm specifically for human movement.
To do so, we create a large and realistic dataset of humans moving in virtual worlds with ground truth optical flow (Fig.~\ref{fig:teaser}(a)).
We train a neural network based on SPyNet \cite{ranjan2016optical} using this dataset and show that it outperforms state of the art optical flow on the test sequences of this dataset  (Fig.~\ref{fig:teaser}(b)).
Furthermore we show that it generalizes to real video sequences  (Fig.~\ref{fig:teaser}(c)).
Here we also extend SPyNet, making it end-to-end trainable.


Several datasets and benchmarks \cite{Geiger2012CVPR,Butler:ECCV:2012,baker2011database} have been established to drive the progress in optical flow.
We argue that these datasets are insufficient for the task of human motion estimation and, despite its importance no attention has been paid to datasets and algorithms for human flow.
One of the main reasons is that dense human motion is extremely difficult to capture accurately in real scenes.
Without ground truth, there has been little work focused specifically on estimating human optical flow.
To advance research on this problem, the community needs a dataset tailored to human flow.


A key observation is that recent work has shown that optical flow methods trained on synthetic data
 \cite{dosovitskiy2015flownet,ranjan2016optical,ilg2016flownet} generalize relatively well to real data.
Additionally, these methods obtain state of the art results with increased realism of the training data \cite{flyingthings,Gaidon:Virtual:CVPR2016}.
This motivates our effort to create a dataset designed for human motion.

To that end, we use the SMPL body model \cite{SMPL:2015} to generate about a hundred thousand different human shapes.
We then place them on random indoor backgrounds and simulate human activities like running, walking, dancing etc. using motion capture data \cite{loper2014mosh}.
Thus, we create a large virtual dataset that captures the statistics of natural human motion. We then train a deep neural network based on spatial pyramids \cite{ranjan2016optical} and evaluate its performance for estimating human motion.
While the dataset can be used to train any flow method, we choose SpyNet because it is compact and computationally efficient.


In summary, our major contributions are: 1) we provide the ``Human Flow dataset'' with 146,020 frame pairs of human bodies in motion with realistic textures and backgrounds; 2) we show that our network outperforms previous optical flow methods by 30\%\/ on the Human Flow dataset, and it generalizes to real world scenes;
3) we extend SPyNet to be fully end-to-end trainable; 4) our neural network is very small (7.8 MB for the network parameters) and runs in real time (32fps), hence it can be potentially used for embedded applications; 5) we provide data, code, and the trained model\footnote{\url{http://github.com/anuragranj/humanflow}} for research purposes.

\section{Related Work}
\textbf{Human Motion.} 
%
%
Human motion can be understood from 2D motion. 
Early work focused on the movement of 2D joint locations \cite{Johansson1973}  or simple motion history images \cite{mhi_davis_2001}.
Optical flow is also a useful cue.
Black et al.~\cite{Black:IEEE:1997} use principal component analysis (PCA) to parametrize human motion but use noisy flow computed from image sequences for training data.
More similar to us, Fablet and Black \cite{Black:ECCV:2002} use a 3D articulated body model and motion capture data to project 3D body motion into 2D optical flow.
They then learn a view-based PCA model of the flow fields.  
We use a more realistic body model to generate a large dataset and use this to train a CNN to directly estimate dense human flow from images.



Only a few works in pose estimation have exploited human motion and, in particular several methods \cite{Fragkiadaki2013,Zuffi:ICCV:2013} use optical flow constraints to improve 2D human pose estimation in videos.
Similar work~\cite{PfisterCZ15,CharlesPMHZ16} propagates pose results temporally using optical flow to encourage time consistency of the estimated bodies. 
Apart from its application in warping between frames, the structural information existing in optical flow has been used for pose estimation alone \cite{flowcap} or in conjunction with an image stream ~\cite{FeichtenhoferPZ16}.


\textbf{Learning Optical Flow.} 
There is a long history of optical flow estimation, which we do not review here.
Instead, we focus on the relatively recent literature on learning flow.
Early work looked at learning flow using Markov Random Fields \cite{Freeman2000}, PCA \cite{wulff2015efficient} , or shallow convolutional models \cite{roth2008learning}.
%
Other methods also combine learning with traditional approaches, formulating flow as a discrete \cite{guney2016ACCV} or continuous \cite{epicflow} optimization problem. 

The most recent methods employ large datasets to estimate optical flow using deep neural networks. Voxel2Voxel \cite{Tran:End2End:2016} is based on volumetric convolutions to predict optical flow using 16 frames simultaneously but does not peform well on benchmarks.
Other methods \cite{dosovitskiy2015flownet,ilg2016flownet,ranjan2016optical} compute two frame optical flow using an end-to-end deep learning approach. FlowNet \cite{dosovitskiy2015flownet} uses the Flying Chairs dataset \cite{dosovitskiy2015flownet} to compute optical flow in an end to end deep network. 
FlowNet 2.0 \cite{ilg2016flownet} uses stacks of networks from FlowNet and performs significantly better, particularly for small motions.
Ranjan and Black \cite{ranjan2016optical} propose a Spatial Pyramid Network that employs a small neural network on each level of an image pyramid to compute optical flow. Their method uses a much smaller number of parameters and achieves similar performance as FlowNet \cite{dosovitskiy2015flownet} using the same training data. 
Since the above methods are not trained with human motions, they do not perform well on our Human Flow dataset.

\textbf{Optical Flow Datasets.} 
Several datasets have been developed to facilitate training and benchmarking of optical flow methods.
Middlebury is limited to small motions \cite{baker2011database}, KITTI is focused on rigid scenes and automotive motions \cite{Geiger2012CVPR}, while Sintel has a limited number of synthetic scenes  \cite{Butler:ECCV:2012}.
%
These datasets are mainly used for evaluation of optical flow methods and are generally too small to support training neural networks.

To learn optical flow using neural networks, more datasets have emerged that contain examples on the order of tens of thousands of frames. 
The Flying Chairs \cite{dosovitskiy2015flownet} dataset contains about 22,000 samples of chairs moving against random backgrounds. 
Although it is not very realistic or diverse, it provides training data for neural networks \cite{dosovitskiy2015flownet,ranjan2016optical} that achieve reasonable results on optical flow benchmarks. 
Even more recent datasets \cite{flyingthings,Gaidon:Virtual:CVPR2016} for optical flow are especially designed for training deep neural networks. 
Flying Things \cite{flyingthings} contains tens of thousands of samples of random 3D objects in motion. 
The Monkaa and Driving scene datasets \cite{flyingthings} contain frames from animated scenes and virtual driving respectively. 
Virtual KITTI \cite{Gaidon:Virtual:CVPR2016} uses graphics to generate scenes like those in KITTI and is two orders of magnitude larger. 
Recent synthetic datasets \cite{Gaidon2014} show that synthetic data can train networks that generalize to real scenes.

For human bodies, the SURREAL dataset \cite{surreal} uses 3D human meshes rendered on top of images to train networks for 2D pose estimation, depth estimation, and body part segmentation. 
While not fully realistic, they show that this data is sufficient to train methods that generalize to real data.
We go beyond their work to address the problem of optical flow.

\section{The Human Flow Dataset}
\begin{figure*}
\centerline{
\includegraphics[width=0.75\textwidth]{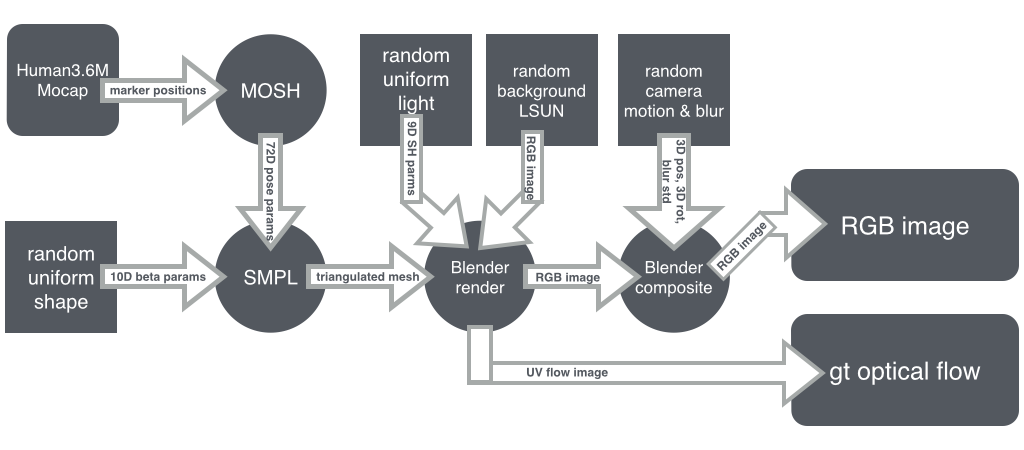}
}
\vspace{-0.5cm}
\caption{Pipeline for generating the RGB frames and ground truth optical flow for the Human Flow dataset.}
\label{fig:pipe}
\vspace{-0.4cm}
\end{figure*}


Our approach generates a realistic dataset of synthetic human motions by simulating them against different realistic backgrounds. As shown in Figure \ref{fig:pipe}, we use the SMPL model \cite{SMPL:2015} to generate a wide variety of different human shapes and appearances. We use Blender \footnote{\url{https://www.blender.org/}} as a rendering engine to generate  synthetic image frames and optical flow. In the rest of the section, we describe each component of our pipeline shown in Figure \ref{fig:pipe}.

\textbf{Body model and Rendering Engine.}
The main challenge in the generation of realistic human optical flow is modeling realistic articulated motions. 
As shown in Figure \ref{fig:pipe}, we use the SMPL model \cite{SMPL:2015}, parameterized by pose and shape parameters to change the body posture and identity. The model also contains a UV appearance map that allows us to change the skin tone, face features and clothing texture of the model.
A key component of Blender in this project is its \emph{Vector pass}. This render pass is typically used for producing motion blur, and it produces the motion in image space of every pixel; i.e. ground truth optical flow. We are mainly interested in the result of this pass, together with the color rendering of the textured bodies.

\textbf{Body Poses.}
In order to obtain a varied set of poses, we use motions from the Human3.6M dataset \cite{h36m_pami}. 
Human3.6M contains five subjects for training (S1, S5, S6, S7, S8) and two for testing (S9, S11). Each subject performs 15 actions twice, resulting in 1,559,985 frames for training and 550,727 for testing. 
These sequences are subsampled at a rate of $16\times$, resulting in 97,499 training and 34,420 testing poses from Human3.6M. The pose data is then converted into SMPL body models using MoSh \cite{loper2014mosh}. We limit each of our pose sequences to 20 frames.

\textbf{Body shapes.}
To maximize the variety of the data, each sequence of $20$ frames uses a random body shape drawn from a uniform distribution of SMPL shape parameters bounded by $[-3,3]$ standard deviations for each shape coefficient according to the shape distribution in CAESAR \cite{robinette2002civilian}. Using a parametric distribution ensures that each sub-sequence of frames has a unique body shape. A uniform distribution has more extreme shapes than the Gaussian distribution inferred originally from CAESAR, while avoiding unlikely shapes by strictly bounding the coefficients.

\textbf{Scene Illumination.}
Optical flow estimation should be robust to different scene illumination. In order to achieve this invariance, we illuminate the bodies with Spherical Harmonics lighting~\cite{spherical_harmonics}. Spherical Harmonics define basis vectors for light directions that are scaled and linearly combined. This compact parameterization is particularly useful for randomizing the scene light. The linear coefficients are randomly sampled with a slight bias towards natural illumination. The coefficients are uniformly sampled between $-0.7$ and $0.7$, apart from the ambient illumination (which is strictly positive and a minimum of $0.3$) and the vertical illumination (which is strictly negative to prevent illumination from below).

\textbf{Body texture.}
To provide a varied set of appearances to the bodies in the scene, we use textures from two different sources. A wide variety of human skin tones is extracted from the CAESAR dataset~\cite{robinette2002civilian}. Given SMPL registrations to CAESAR scans, the original per-vertex color in the CAESAR dataset is transferred into the SMPL texture map. Since fiducial markers were placed on the bodies of CAESAR subjects, we remove them from the textures and inpaint them to produce a natural texture.
The main drawback of CAESAR scans is their homogeneity in terms of outfit, since all of the subjects wore grey shorts and the women wore sports bras. In order to increase the clothing variety, we also use textures extracted from 3D scans. 
A total of 772 textures from 7 different subjects with different clothes were captured. All textures were anonymized by replacing the face by the average face in CAESAR, after correcting it to match the skin tone of the texture.
The datasets were partitioned  $70\% | 30\%$ into training and testing, and each texture dataset was sampled with a $50\%$ chance.

\textbf{Background texture.}
The other crucial component of image appearance is the background. Since human motion rarely happens in front of clean and easily segmentable scenes, realistic backgrounds should be included in the synthetic scenes. We found that using random indoor images from the LSUN dataset~\cite{yu_lsun} as background provided a good compromise between simplicity and the complex task of generating varied full 3D environments.
We used 417,597 images from LSUN categories kitchen, living room, bedroom and dining room.
The background images were placed as billboards 9 meters from the camera, and were not affected by the spherical harmonics lighting.

\textbf{Increasing image realism.}
One of the main differences between synthetic and real images are the imperfections existing in the latter. Fully sharp images rendered with perfectly static virtual cameras do not represent well the images captured in real situations. In order to increase realism, we introduced three types of images imperfections. First, in $30\%$ of the generated images we introduced camera motion between frames. This motion perturbs the location of the camera with Gaussian noise of $1$ centimeter standard deviation, and rotation noise of $0.2$ degrees standard deviation per dimension in an Euler angle representation. Second, motion blur was added to the scene. The motion blur was implemented with the \emph{Vector Blur Node} in Blender, and  integrated over 2 frames sampled with 64 steps between the beginning and end point of the motion.
Finally, general image blur was added to 30\% of the images, as Gaussian blur with a standard
deviation of 1 pixel.

\textbf{Dataset Details.}
In comparison with other optical flow datasets, our dataset is larger by an order of magnitude, containing 135,153 training frames and 10,867 test frames with optical flow ground truth. We keep the resolution small at $256 \times 256$ to facilitate easy deployment for training neural networks. We show the comparisons in Table \ref{tab:evaluation}(a). This also speeds up the rendering process in Blender for generating large amounts of data. Our data is extensive, containing a wide variety of human shapes, poses, actions and virtual backgrounds to support deep learning systems.




\section{Learning}
\begin{figure*}[t]
\centerline{
\includegraphics[width=0.8\linewidth]{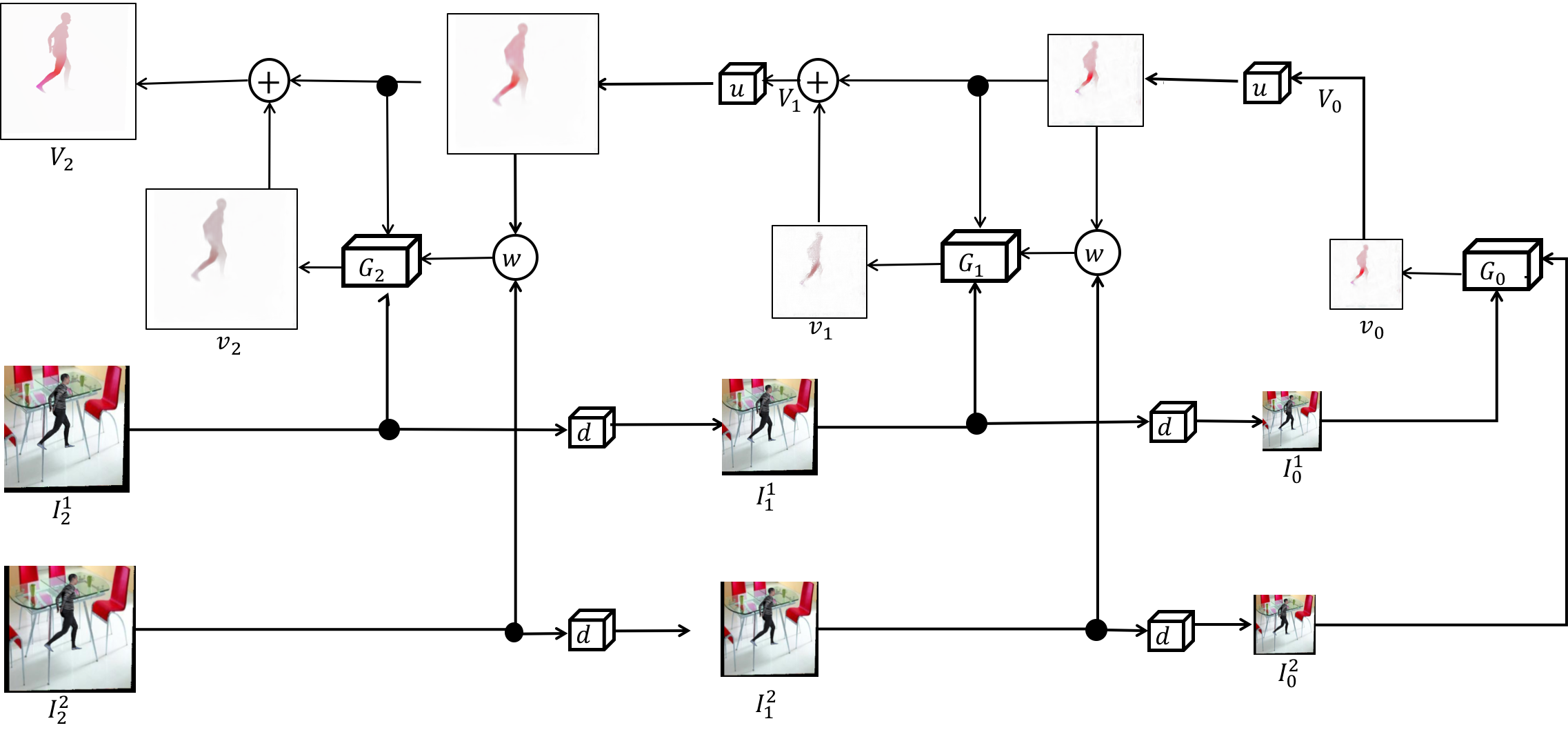}
}
\caption{A Spatial Pyramid Network \cite{ranjan2016optical} for Optical Flow. At each pyramid level, network $G_k$ predicts flow residuals $v_k$ that get added up to produce full flow $V_2$. $w$ is warping operator. $u,d$ are learned convolutional layers that upsample flows and downsample images. The figure shows 3 pyramid levels for simplicity. Our implementation uses 4 levels. }
\label{fig:spynet}
\end{figure*}

Our neural network derives from SPyNet \cite{ranjan2016optical}, which employs different convnets at different levels of an image pyramid. In SPyNet, these convnets are trained independently and sequentially to estimate optical flow. In contrast, we introduce learnable convolutional layers $u,d$, for upsampling and downsampling between pyramid levels. In SPyNet, these are fixed bilinear operators. A differential warping operator, $w$ \cite{jaderberg2015spatial} facilitates our model to be fully differentiable and end-to-end trainable. Thus, we perform joint training of all the convnets at different pyramid levels rather than training them sequentially like SPyNet. We briefly describe the spatial pyramid structure and introduce our network and learning process below.

Our architecture consists of 4 pyramid levels. For simplicity, we show 3 of the 4 pyramid levels in Figure \ref{fig:spynet}. Each level works on a particular resolution of the image. The top level works on the full resolution and the images are downsampled as we move to the bottom of the pyramid. Each level learns a convolutional layer $d$, to perform downsampling of images. Similarly, a convolution layer $u$, is learned for upsampling optical flow. At each level, we also learn a convnet $G_k$ to predict optical flow residuals $v_k$ at that level. These flow residuals get added at each level to produce the full flow, $V_K$ at the finest level of the pyramid.

Each convnet $G_k$ takes a pair of images as inputs along with flow $V_{k-1}$ obtained by upsampling the output of the previous level. The second frame is however warped using $V_{k-1}$ and the triplet $\{I^1_k,  w(I^2_k, V_{k-1}),$ $  V_{k-1}\}$ is fed as input to the convnet $G_k$. The structure of the convnets $G_k$ is same as SPyNet. Each of the convnets $G_k$ is a five layer network containing \{32, 64, 32, 16, 2\} feature maps with 7x7 kernels.
At each level, the downsampling layers $d$ learn 3x3 convolutional kernels with 6 feature maps to operate on 6 channels of image pairs. Similarly, the upsampling layers $u$ learn 4x4 convolutional kernels with 2 feature maps to operate on 2-channel flows.  We use $w$ to refer to a bilinear warping operator which is non-learnable. The general structure of spatial pyramids can be seen in \cite{ranjan2016optical}.
We import the weights of our convnets $G_i$ from the first four convnets $\{G_0,G_1,G_2,G_3\}$ of SPyNet pre-trained on the Flying Chairs dataset \cite{dosovitskiy2015flownet}. We use a differentiable warping operator, $w$ \cite{jaderberg2015spatial}. We now construct our fully differentiable spatial pyramid architecture and train it end-to-end minimizing an End Point Error (EPE).


\textbf{Hyperparameters.} We use Adam \cite{kingma2014adam} to optimize our loss at a constant learning rate of $10^{-6}$, $\beta_1=0.9$ and $\beta_2=0.999$. We use a batch size of 8 and run 4000 iterations per epoch. We train our model for 100 epochs on the Human Flow dataset. We use the Torch7\footnote{\url{http://torch.ch}} framework for our implementation and use four Nvidia K80 GPUs to train in parallel. It takes 1 day for our model to train.

\textbf{Data Augmentations.} We also augment our data by applying several transformations and adding noise. Although our dataset is quite large, augmentation improves the quality of results on real scenes. In particular, we apply scaling in the range of $[0.3, 3]$, and rotations in $[-17^{\circ}, 17^{\circ}]$. We randomly crop the images and flows to $256 \times 256$ at the finest level. We sample uniform Gaussian noise $\mathcal{N}(0,1)$ and add it to the images with a weight factor of 1:10. We also apply color jitter with additive brightness, contrast, and saturation changes. The dataset is normalized to have zero mean and unit standard deviation using \cite{he2015deep}.

\section{Experiments}
\begin{figure*}[t]
  \centering
    \includegraphics[width=\textwidth]{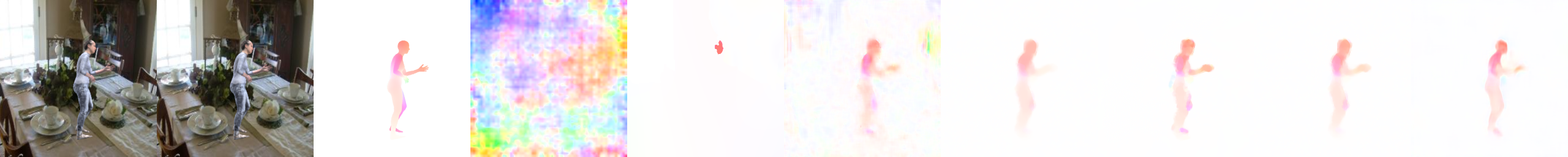}
    \includegraphics[width=\textwidth]{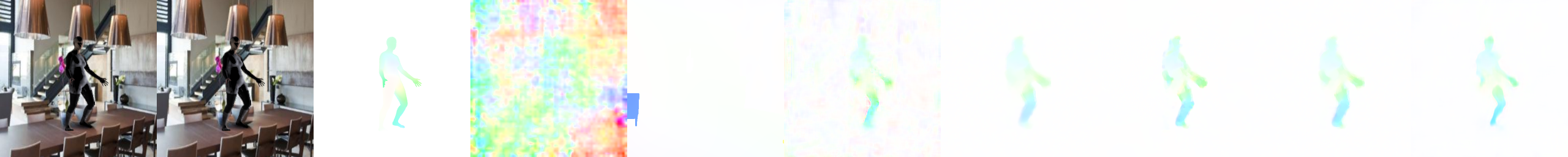}
    \includegraphics[width=\textwidth]{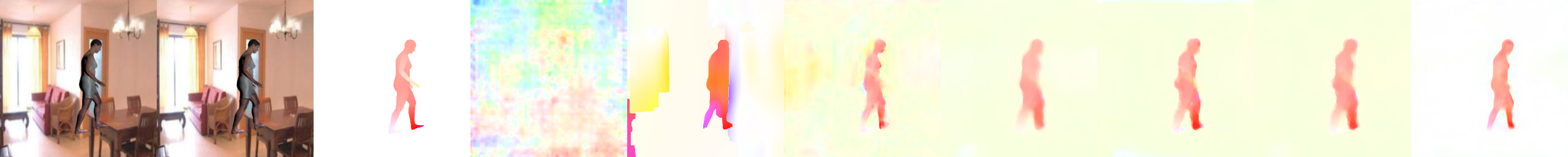}
    \includegraphics[width=\textwidth]{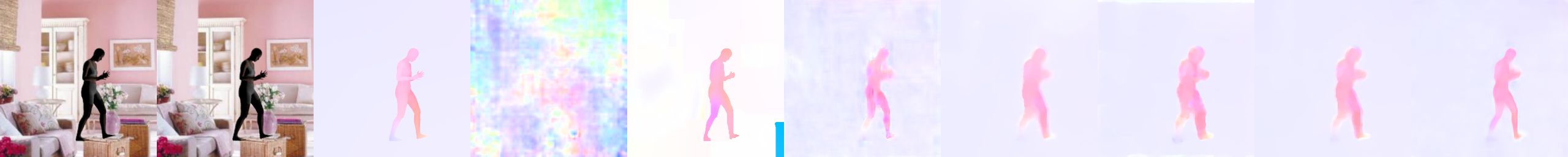}
    \includegraphics[width=\textwidth]{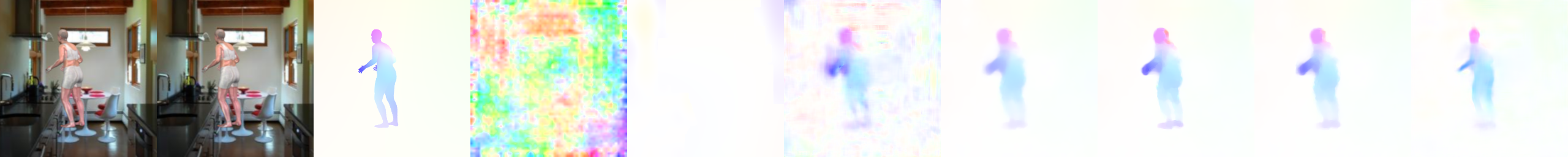}

\begin{flushleft}
\vspace{-0.2in}
\hspace{0.07in} {\scriptsize Frame 1} \hspace{0.05in} {\scriptsize Frame 2}\hspace{0.07in} {\scriptsize Ground Truth}\hspace{0.05in} {\scriptsize FlowNetS}\hspace{0.034in} {\scriptsize PCA-Layers} \hspace{0.070in}{\scriptsize SPyNet}\hspace{0.17in} {\scriptsize EpicFlow} \hspace{0.1in} {\scriptsize LDOF}\hspace{0.1in} {\scriptsize FlowFields}\hspace{0.05in}{\scriptsize HumanFlow}\end{flushleft}
\vspace{-0.1in}
\caption{Visual comparison of optical flow estimates using different methods on the Human Flow test set. From left to right, we show Frame 1, Frame 2, Ground Truth flow, results on FlowNetS \cite{dosovitskiy2015flownet}, PCA-Layers \cite{wulff2015efficient}, SPyNet \cite{ranjan2016optical}, EpicFlow \cite{epicflow}, LDOF \cite{brox2009large}, FlowFields \cite{flowfields} and HumanFlow (ours).
}
\label{fig:results}
\end{figure*}

We compare the average End Point Errors (EPEs) of competing methods in Table \ref{tab:evaluation}(b) along with the time for evaluation. Human motion is complex and general optical flow methods fail to capture it. Our trained network outperforms previous methods, and SPyNet \cite{ranjan2016optical} in particular, in terms of average EPE on the Human Flow Dataset. This indicates that other optical flow networks can also improve their performance on human motion using our dataset. We show visual comparisons in Figure \ref{fig:results}.

Optical flow methods that employ learning using large datasets such as FlowNetS \cite{dosovitskiy2015flownet} show poor generalization on our dataset. Since the results of FlowNetS are very close to the zero flow baseline (Table \ref{tab:evaluation}), we cross-verify by evaluating FlowNetS on a mixture of Flying Chairs \cite{dosovitskiy2015flownet} and Human Flow dataset and observe that the flow outputs on human flow dataset is quite random (see Figure \ref{fig:results}). The main reason is that our dataset contains a significant amount of small motions and it is known that FlowNetS does not perform very well on small motions. SPyNet \cite{ranjan2016optical} however performs quite well and is able to generalize to body motions. The results however look noisy in many cases.

Our dataset employs a layered structure where a human is placed against a background. As such layered methods like PCA-layers \cite{wulff2015efficient} perform very well on a few images (row 4 in Figure \ref{fig:results}) where they are able to segment a person from the background. However, in most cases, they do not obtain good segmentation into layers.

Previous state-of-the-art methods like LDOF \cite{brox2009large}, EpicFlow \cite{epicflow} and FlowFields \cite{flowfields} perform much better than others. They get a good overall shape, and smooth backgrounds. However, their estimation is quite blurred. They tend to miss the sharp edges that are typical of human hands and legs. They are also significantly slower.

In contrast, our network outperforms the state of the art optical flow methods by 30\% on the Human Flow dataset. Our qualitative results show that our method can capture sharp details like hands and legs of the person.
Since, our test data is comparatively large, we evaluate only the fastest optical methods on our dataset.

\begin{table}[t]
\begin{tabular}{lccc}
\hline
Dataset          &\begin{tabular}[c]{@{}l@{}}{\small \# Train}\\ {\small Frames}\end{tabular} & \begin{tabular}[c]{@{}l@{}}{\small \# Test} \\ {\small Frames}\end{tabular} & {\small Resolution} \\ \hline
{\small MPI Sintel\cite{Butler:ECCV:2012}} & {\small $1,064  $}    &{\small $564 $ }    & {\small $1024 \times 436$ }\\
{\small KITTI 2012\cite{Geiger2012CVPR}}      & {\small $194   $  }  & {\small $195 $ }    & {\small $1226 \times 370$} \\
{\small KITTI 2015\cite{menze2015object}} & {\small $200 $ }   &{\small  $200 $ }    & {\small $1242 \times 375$} \\
{\small Virtual Kitti\cite{Gaidon:Virtual:CVPR2016}}   & {\small $21,260 $ }   & $-   $    &{\small $1242 \times 375 $} \\
{\small Flying Chairs\cite{dosovitskiy2015flownet}}   & {\small $22,232 $ }   &{\small $640 $  }   &{\small $512 \times 384 $} \\
{\small Flying Things\cite{flyingthings} }& {\small $21,818 $}    & {\small $4,248$ }    & {\small $960 \times 540 $} \\
{\small Monkaa\cite{flyingthings} }          & {\small $8,591  $ }   & $-   $     &{\small $960 \times 540 $} \\
{\small Driving\cite{flyingthings}}         & {\small $4,392  $}    & $-   $     & {\small $960 \times 540 $} \\
{\small Human Flow}       & {\small $135,153$ }   & {\small $ 10,867$}     & {\small $256 \times 256 $ }\\ \hline
\multicolumn{4}{c}{(a)}
\end{tabular}
\quad
\begin{tabular}{lcc}
\hline
Method     & {\small AEPE} & {\small Time(s)} \\ \hline
{\small Zero} & {\small 0.6611} &  - \\ \hline
{\small FlowNet\cite{dosovitskiy2015flownet} }& {\small 0.5846}  &  {\small 0.080} \\
{\small PCA Flow\cite{wulff2015efficient}} &{\small 0.3652}  & {\small 10.357}        \\
{\small SPyNet\cite{ranjan2016optical} } &{\small 0.2066}   &{\small 0.038}    \\
{\small Epic Flow\cite{epicflow}} &{\small 0.1940}    & {\small 1.863} \\
{\small LDOF\cite{brox2009large}} & {\small 0.1881}   &  {\small 8.620}       \\
{\small Flow Fields\cite{flowfields} }& {\small 0.1709}  &{\small 4.204}   \\
{\small HumanFlow} &{\small \textbf{0.1164}}  &	{\small \textbf{0.031}}	\\ \hline
\multicolumn{3}{c}{(b)}
\end{tabular}
\vspace{0.02in}
\caption{(a) Comparison of Human Flow dataset with previous optical flow datasets. (b) EPE comparisons and evaluation times of different optical flow methods on the Human Flow dataset. Zero refers to the EPE when zero flow is always used for evaluation.}
\label{tab:evaluation}
\end{table}

\begin{figure}
\centering     
\subfigure[]{\label{fig:persondetect}\includegraphics[width=60mm]{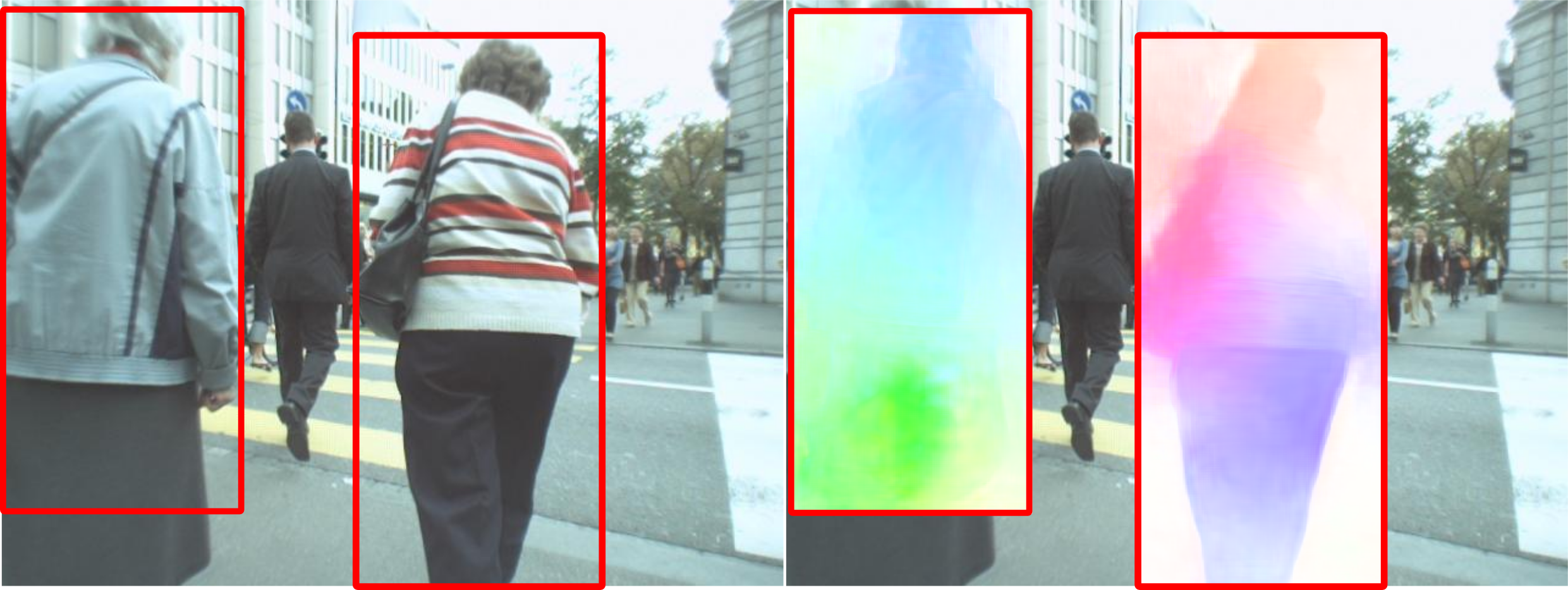}}
\subfigure[]{\label{fig:realpeople}\includegraphics[width=60mm]{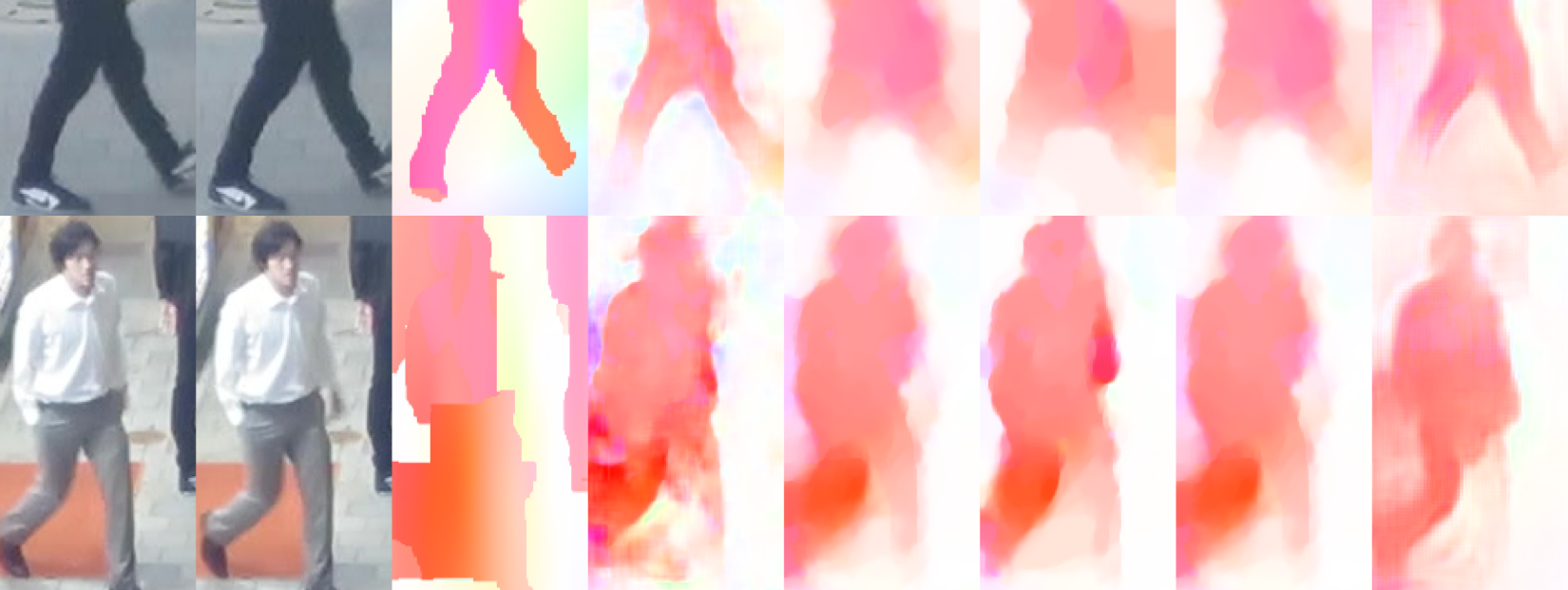}}
\caption{(a) We use a person detector to crop out people from real-world scenes (left) and use our model to compute optical flow on the cropped section (right).  (b) Visual comparison on real scenes. From left to right, we show Frame 1, Frame 2, results on PCA-Layers \cite{wulff2015efficient}, and SPyNet \cite{ranjan2016optical}, EpicFlow \cite{epicflow}, LDOF \cite{brox2009large}, FlowFields \cite{flowfields} and HumanFlow (ours). Our method is good at picking up limb and head motion.}
\end{figure}

\textbf{Real Scenes.}
We show a visual comparison of results on real-world scenes of people in motion. We collect these scenes by cropping people from real world videos as shown in Figure \ref{fig:persondetect}. We use DPM \cite{Felzenszwalb2010PAMI} for detecting people and compute bounding box regions in two frames using the ground truth of the MOT16 dataset \cite{MilanL0RS16}.

We visually compare our results with other popular optical flow methods on real world scenes in Figure \ref{fig:realpeople}. The performance of PCA-Layers \cite{wulff2015efficient} is highly dependent on its ability to segment. Hence, we see only a few cases where it looks visually correct. SPyNet \cite{ranjan2016optical} gets the overall shape but the results look noisy in certain image parts. While LDOF \cite{brox2009large}, EpicFlow \cite{epicflow} and FlowFields \cite{flowfields} generally perform  well, they often find it difficult to resolve the legs, hands and head of the person. The results from HumanFlow look appealing especially while resolving the overall human shape, and various parts like legs, hands and the human head. HumanFlow also performs very well under occlusions when the full body is not visible.

\textbf{Timing Evaluation.}
Although some of the methods do quite well on our benchmark, they tend to be slow due to their complex nature. As such, they are not likely to be used in real time or embedded applications. We show timing comparisons on a pair of frames in Table \ref{tab:evaluation}. We show that learning with human flow data can make our model a lot simpler while simultaneously making it fast and accurate for capturing human motion. Our model takes 31 ms for inference on NVIDIA TitanX. As such it can run in real time at 32 fps.

\textbf{Network Size.}
Employing spatial pyramid structures \cite{ranjan2016optical} to neural network architectures leads to significant reduction of size. As such our network is quite small and can be stored in 7.8 MB of memory. Our network has a total of 4.2 million learnable parameters making it fast and easy to train.

\section{Conclusion and Future Work}
In summary, we created an extensive dataset containing images of realistic human shapes in motion together with groundtruth optical flow.
The realism and extent of this dataset, together with an end-to-end trained system, allows our new  HumanFlow method to outperform
previous state-of-the-art optical flow methods in our new human-specific dataset.
Furthermore, we show that our method generalizes to human motion in real world scenes for optical flow computation.
Our method is compact and runs in real time making it highly suitable for phones and embedded devices.

In future work, we plan to model more subtle human motions such as those of faces and hands. 
We also plan to generate training sequences contining multiple, interacting people, and more complex 3D scene motions.
Additionally, we plan to add 3D clothing and accessories.
The dataset and our focus on human flow opens up a number of research directions in human motion understanding and optical flow computation.
We would like to extend our dataset by modeling more diverse clothing and outdoor scenarios.
A direction of potentially high impact for this work is to integrate it in end-to-end systems for action recognition,
which typically take pre-computed optical flow as input.
The real-time nature of the method could support motion-based interfaces, potentially even on devices like cell phones with limited computing power.
The data, model, and training code is available, enabling researchers to apply this to problems involving human motion.


\section*{Acknowledgements}
We thank Siyu Tang for compiling the person detections and Yiyi Liao for helping us with
optical flow evaluation.  We thank Cristian Sminchisescu for the
Human3.6M mocap marker data.
\bibliography{egbib}

\begin{thebibliography}{44}
\providecommand{\natexlab}[1]{#1}
\providecommand{\url}[1]{\texttt{#1}}
\expandafter\ifx\csname urlstyle\endcsname\relax
  \providecommand{\doi}[1]{doi: #1}\else
  \providecommand{\doi}{doi: \begingroup \urlstyle{rm}\Url}\fi

\bibitem[Bailer et~al.(2015)Bailer, Taetz, and Stricker]{flowfields}
Christian Bailer, Bertram Taetz, and Didier Stricker.
\newblock Flow fields: Dense correspondence fields for highly accurate large
  displacement optical flow estimation.
\newblock In \emph{Proceedings of the IEEE International Conference on Computer
  Vision}, pages 4015--4023, 2015.

\bibitem[Baker et~al.(2011)Baker, Scharstein, Lewis, Roth, Black, and
  Szeliski]{baker2011database}
Simon Baker, Daniel Scharstein, JP~Lewis, Stefan Roth, Michael~J Black, and
  Richard Szeliski.
\newblock A database and evaluation methodology for optical flow.
\newblock \emph{International Journal of Computer Vision}, 92\penalty0
  (1):\penalty0 1--31, 2011.

\bibitem[Black et~al.(1997)Black, Yacoob, Jepson, and Fleet]{Black:IEEE:1997}
M.~J. Black, Y.~Yacoob, A.~D. Jepson, and D.~J. Fleet.
\newblock Learning parameterized models of image motion.
\newblock In \emph{IEEE Conf. on Computer Vision and Pattern Recognition,
  CVPR-97}, pages 561--567, Puerto Rico, June 1997.

\bibitem[Brox et~al.(2009)Brox, Bregler, and Malik]{brox2009large}
Thomas Brox, Christoph Bregler, and Jitendra Malik.
\newblock Large displacement optical flow.
\newblock In \emph{Computer Vision and Pattern Recognition, 2009. CVPR 2009.
  IEEE Conference on}, pages 41--48. IEEE, 2009.

\bibitem[Butler et~al.(2012)Butler, Wulff, Stanley, and
  Black]{Butler:ECCV:2012}
D.~J. Butler, J.~Wulff, G.~B. Stanley, and M.~J. Black.
\newblock A naturalistic open source movie for optical flow evaluation.
\newblock In {A. Fitzgibbon et al. (Eds.)}, editor, \emph{European Conf. on
  Computer Vision (ECCV)}, Part IV, LNCS 7577, pages 611--625. Springer-Verlag,
  October 2012.

\bibitem[Charles et~al.(2016)Charles, Pfister, Magee, Hogg, and
  Zisserman]{CharlesPMHZ16}
James Charles, Tomas Pfister, Derek~R. Magee, David~C. Hogg, and Andrew
  Zisserman.
\newblock Personalizing human video pose estimation.
\newblock In \emph{CVPR}, pages 3063--3072. IEEE Computer Society, 2016.
\newblock ISBN 978-1-4673-8851-1.
\newblock URL
  \url{http://ieeexplore.ieee.org/xpl/mostRecentIssue.jsp?punumber=7776647}.

\bibitem[D. et~al.(2008)D., Roth, Lewis, and Black]{roth2008learning}
Sun D., S~Roth, JP~Lewis, and MJ~Black.
\newblock Learning optical flow.
\newblock In \emph{ECCV}, pages 83--97, 2008.

\bibitem[Davis(2001)]{mhi_davis_2001}
J.~W. Davis.
\newblock Hierarchical motion history images for recognizing human motion.
\newblock In \emph{Detection and Recognition of Events in Video}, pages 39--46,
  2001.
\newblock URL \url{http://dx.doi.org/10.1109/EVENT.2001.938864}.

\bibitem[Dosovitskiy et~al.(2015)Dosovitskiy, Fischery, Ilg, Hazirbas, Golkov,
  van~der Smagt, Cremers, Brox, et~al.]{dosovitskiy2015flownet}
Alexey Dosovitskiy, Philipp Fischery, Eddy Ilg, Caner Hazirbas, Vladimir
  Golkov, Patrick van~der Smagt, Daniel Cremers, Thomas Brox, et~al.
\newblock Flownet: Learning optical flow with convolutional networks.
\newblock In \emph{2015 IEEE International Conference on Computer Vision
  (ICCV)}, pages 2758--2766. IEEE, 2015.

\bibitem[Fablet and Black(2002)]{Black:ECCV:2002}
R.~Fablet and M.~J. Black.
\newblock Automatic detection and tracking of human motion with a view-based
  representation.
\newblock In \emph{European Conf. on Computer Vision, ECCV 2002}, volume~1 of
  \emph{LNCS 2353}, pages 476--491. Springer-Verlag, 2002.

\bibitem[Feichtenhofer et~al.(2016)Feichtenhofer, Pinz, and
  Zisserman]{FeichtenhoferPZ16}
Christoph Feichtenhofer, Axel Pinz, and Andrew Zisserman.
\newblock Convolutional two-stream network fusion for video action recognition.
\newblock In \emph{CVPR}, pages 1933--1941. IEEE Computer Society, 2016.
\newblock ISBN 978-1-4673-8851-1.
\newblock URL
  \url{http://ieeexplore.ieee.org/xpl/mostRecentIssue.jsp?punumber=7776647}.

\bibitem[Felzenszwalb et~al.(2010)Felzenszwalb, Girshick, McAllester, and
  Ramanan]{Felzenszwalb2010PAMI}
P.~F. Felzenszwalb, R.~B. Girshick, D.~McAllester, and D.~Ramanan.
\newblock Object detection with discriminatively trained part-based models.
\newblock \emph{TPAMI}, 32\penalty0 (9):\penalty0 1627--1645, 2010.

\bibitem[Fragkiadaki et~al.(2013)Fragkiadaki, Hu, and Shi]{Fragkiadaki2013}
K.~Fragkiadaki, H.~Hu, and J.~Shi.
\newblock Pose from flow and flow from pose.
\newblock In \emph{2013 IEEE Conference on Computer Vision and Pattern
  Recognition}, pages 2059--2066, June 2013.
\newblock \doi{10.1109/CVPR.2013.268}.

\bibitem[Freeman et~al.(2000)Freeman, Pasztor, and Carmichael]{Freeman2000}
William~T. Freeman, Egon~C. Pasztor, and Owen~T. Carmichael.
\newblock Learning low-level vision.
\newblock \emph{International Journal of Computer Vision}, 40\penalty0
  (1):\penalty0 25--47, 2000.
\newblock ISSN 1573-1405.
\newblock \doi{10.1023/A:1026501619075}.
\newblock URL \url{http://dx.doi.org/10.1023/A:1026501619075}.

\bibitem[Gaidon et~al.(2016)Gaidon, Wang, Cabon, and
  Vig]{Gaidon:Virtual:CVPR2016}
A~Gaidon, Q~Wang, Y~Cabon, and E~Vig.
\newblock Virtual worlds as proxy for multi-object tracking analysis.
\newblock In \emph{CVPR}, 2016.

\bibitem[Gaidon et~al.(2014)Gaidon, Harchaoui, and Schmid]{Gaidon2014}
Adrien Gaidon, Zaid Harchaoui, and Cordelia Schmid.
\newblock Activity representation with motion hierarchies.
\newblock \emph{International Journal of Computer Vision}, 107\penalty0
  (3):\penalty0 219--238, 2014.
\newblock ISSN 1573-1405.
\newblock \doi{10.1007/s11263-013-0677-1}.

\bibitem[Geiger et~al.(2012)Geiger, Lenz, and Urtasun]{Geiger2012CVPR}
Andreas Geiger, Philip Lenz, and Raquel Urtasun.
\newblock Are we ready for autonomous driving? the {KITTI} vision benchmark
  suite.
\newblock In \emph{Conference on Computer Vision and Pattern Recognition
  (CVPR)}, 2012.

\bibitem[Geman and Geman(2016)]{scienceselfies}
Donald Geman and Stuart Geman.
\newblock Opinion: Science in the age of selfies.
\newblock \emph{Proceedings of the National Academy of Sciences}, 113\penalty0
  (34):\penalty0 9384--9387, 2016.

\bibitem[Green(2003)]{spherical_harmonics}
Robin Green.
\newblock {Spherical Harmonic Lighting: The Gritty Details}.
\newblock \emph{Archives of the Game Developers Conference}, March 2003.
\newblock URL
  \url{http://www.research.scea.com/gdc2003/spherical-harmonic-lighting.pdf}.

\bibitem[G{\"u}ney and Geiger(2016)]{guney2016ACCV}
Fatma G{\"u}ney and Andreas Geiger.
\newblock Deep discrete flow.
\newblock In \emph{Asian Conference on Computer Vision (ACCV)}, 2016.

\bibitem[He et~al.(2015)He, Zhang, Ren, and Sun]{he2015deep}
Kaiming He, Xiangyu Zhang, Shaoqing Ren, and Jian Sun.
\newblock Deep residual learning for image recognition.
\newblock \emph{arXiv preprint arXiv:1512.03385}, 2015.

\bibitem[Ilg et~al.(2016)Ilg, Mayer, Saikia, Keuper, Dosovitskiy, and
  Brox]{ilg2016flownet}
Eddy Ilg, Nikolaus Mayer, Tonmoy Saikia, Margret Keuper, Alexey Dosovitskiy,
  and Thomas Brox.
\newblock Flownet 2.0: Evolution of optical flow estimation with deep networks.
\newblock \emph{arXiv preprint arXiv:1612.01925}, 2016.

\bibitem[Ionescu et~al.(2014)Ionescu, Papava, Olaru, and
  Sminchisescu]{h36m_pami}
Catalin Ionescu, Dragos Papava, Vlad Olaru, and Cristian Sminchisescu.
\newblock Human3.6m: Large scale datasets and predictive methods for 3d human
  sensing in natural environments.
\newblock \emph{IEEE Transactions on Pattern Analysis and Machine
  Intelligence}, 36\penalty0 (7):\penalty0 1325--1339, jul 2014.

\bibitem[Jaderberg et~al.(2015)Jaderberg, Simonyan, Zisserman,
  et~al.]{jaderberg2015spatial}
Max Jaderberg, Karen Simonyan, Andrew Zisserman, et~al.
\newblock Spatial transformer networks.
\newblock In \emph{Advances in Neural Information Processing Systems}, pages
  2017--2025, 2015.

\bibitem[Jhuang et~al.(2013)Jhuang, Gall, Zuffi, Schmid, and
  Black]{Jhuang:ICCV:2013}
Hueihan Jhuang, Juergen Gall, Silvia Zuffi, Cordelia Schmid, and Michael~J.
  Black.
\newblock Towards understanding action recognition.
\newblock In \emph{IEEE International Conference on Computer Vision (ICCV)},
  pages 3192--3199, Sydney, Australia, December 2013. IEEE.

\bibitem[Johansson(1973)]{Johansson1973}
Gunnar Johansson.
\newblock Visual perception of biological motion and a model for its analysis.
\newblock \emph{Perception {\&} Psychophysics}, 14\penalty0 (2):\penalty0
  201--211, 1973.
\newblock ISSN 1532-5962.
\newblock \doi{10.3758/BF03212378}.

\bibitem[Kingma and Ba(2014)]{kingma2014adam}
Diederik Kingma and Jimmy Ba.
\newblock Adam: A method for stochastic optimization.
\newblock \emph{arXiv preprint arXiv:1412.6980}, 2014.

\bibitem[Kuehne et~al.(2011)Kuehne, Jhuang, Garrote, Poggio, and
  Serre]{kuehne2011hmdb}
Hildegard Kuehne, Hueihan Jhuang, Est{\'\i}baliz Garrote, Tomaso Poggio, and
  Thomas Serre.
\newblock Hmdb: a large video database for human motion recognition.
\newblock In \emph{Computer Vision (ICCV), 2011 IEEE International Conference
  on}, pages 2556--2563. IEEE, 2011.

\bibitem[Loper et~al.(2014)Loper, Mahmood, and Black]{loper2014mosh}
Matthew Loper, Naureen Mahmood, and Michael~J Black.
\newblock Mosh: Motion and shape capture from sparse markers.
\newblock \emph{ACM Transactions on Graphics (TOG)}, 33\penalty0 (6):\penalty0
  220, 2014.

\bibitem[Loper et~al.(2015)Loper, Mahmood, Romero, Pons-Moll, and
  Black]{SMPL:2015}
Matthew Loper, Naureen Mahmood, Javier Romero, Gerard Pons-Moll, and Michael~J.
  Black.
\newblock {SMPL}: A skinned multi-person linear model.
\newblock \emph{ACM Trans. Graphics (Proc. SIGGRAPH Asia)}, 34\penalty0
  (6):\penalty0 248:1--248:16, October 2015.

\bibitem[Menze and Geiger(2015)]{menze2015object}
Moritz Menze and Andreas Geiger.
\newblock Object scene flow for autonomous vehicles.
\newblock In \emph{Proceedings of the IEEE Conference on Computer Vision and
  Pattern Recognition}, pages 3061--3070, 2015.

\bibitem[Milan et~al.(2016)Milan, Leal{-}Taix{\'{e}}, Reid, Roth, and
  Schindler]{MilanL0RS16}
Anton Milan, Laura Leal{-}Taix{\'{e}}, Ian~D. Reid, Stefan Roth, and Konrad
  Schindler.
\newblock {MOT16:} {A} benchmark for multi-object tracking.
\newblock \emph{arXiv:1603.00831}, 2016.

\bibitem[N.Mayer et~al.(2016)N.Mayer, E.Ilg, P.H{\"a}usser, P.Fischer,
  D.Cremers, A.Dosovitskiy, and T.Brox]{flyingthings}
N.Mayer, E.Ilg, P.H{\"a}usser, P.Fischer, D.Cremers, A.Dosovitskiy, and T.Brox.
\newblock A large dataset to train convolutional networks for disparity,
  optical flow, and scene flow estimation.
\newblock In \emph{IEEE International Conference on Computer Vision and Pattern
  Recognition (CVPR)}, 2016.
\newblock URL
  \url{http://lmb.informatik.uni-freiburg.de/Publications/2016/MIFDB16}.
\newblock arXiv:1512.02134.

\bibitem[Pfister et~al.(2015)Pfister, Charles, and Zisserman]{PfisterCZ15}
Tomas Pfister, James Charles, and Andrew Zisserman.
\newblock Flowing convnets for human pose estimation in videos.
\newblock In \emph{ICCV}, pages 1913--1921. IEEE Computer Society, 2015.
\newblock ISBN 978-1-4673-8391-2.
\newblock URL
  \url{http://ieeexplore.ieee.org/xpl/mostRecentIssue.jsp?punumber=7407725}.

\bibitem[Ranjan and Black(2016)]{ranjan2016optical}
Anurag Ranjan and Michael~J Black.
\newblock Optical flow estimation using a spatial pyramid network.
\newblock \emph{arXiv preprint arXiv:1611.00850}, 2016.

\bibitem[Revaud et~al.(2015)Revaud, Weinzaepfel, Harchaoui, and
  Schmid]{epicflow}
Jerome Revaud, Philippe Weinzaepfel, Zaid Harchaoui, and Cordelia Schmid.
\newblock {EpicFlow: Edge-Preserving Interpolation of Correspondences for
  Optical Flow}.
\newblock In \emph{{Computer Vision and Pattern Recognition}}, 2015.

\bibitem[Robinette et~al.(2002)Robinette, Blackwell, Daanen, Boehmer, and
  Fleming]{robinette2002civilian}
Kathleen~M Robinette, Sherri Blackwell, Hein Daanen, Mark Boehmer, and Scott
  Fleming.
\newblock Civilian american and european surface anthropometry resource
  (caesar), final report. volume 1. summary.
\newblock Technical report, DTIC Document, 2002.

\bibitem[Romero et~al.(2015)Romero, Loper, and Black]{flowcap}
Javier Romero, Matthew Loper, and Michael~J. Black.
\newblock {FlowCap}: {2D} human pose from optical flow.
\newblock In \emph{Pattern Recognition, Proc. 37th German Conference on Pattern
  Recognition (GCPR)}, volume LNCS 9358, pages 412--423. Springer, 2015.

\bibitem[Soomro et~al.(2012)Soomro, Zamir, and Shah]{soomro2012ucf101}
Khurram Soomro, Amir~Roshan Zamir, and Mubarak Shah.
\newblock Ucf101: A dataset of 101 human actions classes from videos in the
  wild.
\newblock \emph{arXiv preprint arXiv:1212.0402}, 2012.

\bibitem[Tran et~al.(2016)Tran, Bourdev, Fergus, Torresani, and
  Paluri]{Tran:End2End:2016}
Du~Tran, Lubomir Bourdev, Rob Fergus, Lorenzo Torresani, and Manohar Paluri.
\newblock Deep {End2End} {Voxel2Voxel} prediction.
\newblock In \emph{The 3rd Workshop on Deep Learning in Computer Vision}, 2016.

\bibitem[Varol et~al.()Varol, Romero, Martin, Mahmood, Black, Laptev, and
  Schmid]{surreal}
G\"{u}l Varol, Javier Romero, Xavier Martin, Naureen Mahmood, Michael Black,
  Ivan Laptev, and Cordelia Schmid.
\newblock Learning from synthetic humans.
\newblock In \emph{CVPR}.

\bibitem[Wulff and Black(2015)]{wulff2015efficient}
Jonas Wulff and Michael~J Black.
\newblock Efficient sparse-to-dense optical flow estimation using a learned
  basis and layers.
\newblock In \emph{2015 IEEE Conference on Computer Vision and Pattern
  Recognition (CVPR)}, pages 120--130. IEEE, 2015.

\bibitem[Yu et~al.(2015)Yu, Zhang, Song, Seff, and Xiao]{yu_lsun}
Fisher Yu, Yinda Zhang, Shuran Song, Ari Seff, and Jianxiong Xiao.
\newblock Lsun: Construction of a large-scale image dataset using deep learning
  with humans in the loop.
\newblock \emph{arXiv:1506.03365}, 2015.

\bibitem[Zuffi et~al.(2013)Zuffi, Romero, Schmid, and Black]{Zuffi:ICCV:2013}
Silvia Zuffi, Javier Romero, Cordelia Schmid, and Michael~J Black.
\newblock Estimating human pose with flowing puppets.
\newblock In \emph{IEEE International Conference on Computer Vision (ICCV)},
  pages 3312--3319, 2013.

\end{thebibliography}
\end{document}